\documentclass[letterpaper, 10 pt, conference]{ieeeconf}  % Comment this line out if you need a4paper

\IEEEoverridecommandlockouts                              % This command is only needed if 
\usepackage{amsmath} % assumes amsmath package installed
\usepackage{graphicx}
\usepackage{silence}
\usepackage{subcaption}
\usepackage{hyperref}

\title{\Large \bf Real-Time Object Pose Estimation with Pose Interpreter Networks}
\author{Jimmy Wu$^{1}$, Bolei Zhou$^{1}$, Rebecca Russell$^{2}$, Vincent Kee$^{2}$, Syler Wagner$^{3}$, Mitchell Hebert$^{2}$, \\Antonio Torralba$^{1}$, and David M.S. Johnson$^{3}$\\
\thanks{$^{1}$JW, BZ, and AT are with the MIT Computer Science and Artificial Intelligence Laboratory, Cambridge, MA, USA}
\thanks{$^{2}$RR, VK, MH are with the Charles Stark Draper Laboratory, Cambridge, MA, USA}
\thanks{$^{3}$SW and DMSJ are with Dexai Robotics, Boston, MA, USA}
\thanks{Corresponding authors: Jimmy Wu \texttt{jimmywu@alum.mit.edu} and David M.S. Johnson \texttt{dave@dexai.com}}
\thanks{Datasets, code, and pretrained models are available at \url{https://github.com/jimmyyhwu/pose-interpreter-networks}
}
}

\begin{document}

\maketitle
\thispagestyle{empty}
\pagestyle{empty}

%%%%%%%%%%%%%%%%%%%%%%%%%%%%%%%%%%%%%%%%%%%%%%%%%%%%%%%%%%%%%%%%%%%%%%%%%%%%%%%%
\begin{abstract}

In this work, we introduce pose interpreter networks for 6-DoF object pose estimation. In contrast to other CNN-based approaches to pose estimation that require expensively annotated object pose data, our pose interpreter network is trained entirely on synthetic pose data. We use object masks as an intermediate representation to bridge real and synthetic. We show that when combined with a segmentation model trained on RGB images, our synthetically trained pose interpreter network is able to generalize to real data. Our end-to-end system for object pose estimation runs in real-time (20 Hz) on live RGB data, without using depth information or ICP refinement.

\end{abstract}

%%%%%%%%%%%%%%%%%%%%%%%%%%%%%%%%%%%%%%%%%%%%%%%%%%%%%%%%%%%%%%%%%%%%%%%%%%%%%%%%
\section{Introduction}

Object pose estimation is an important task relevant to many applications in robotics, such as robotic object manipulation and warehouse automation. In the past, 6-DoF object pose estimation has been tackled using template matching between 3D models and images \cite{rothganger20063d}, which uses local features such as SIFT \cite{lowe1999object} to recover the pose of highly textured objects. Recently, there has been growing interest in object manipulation as a result of the Amazon Picking Challenge \cite{correll2016analysis}, leading to the introduction of a number of different approaches for 6-DoF object pose estimation, specifically in the competition setting \cite{zeng2017multi,hernandez2016team,jonschkowski2016probabilistic,schwarz2017nimbro,schwarz2016rgb}. Many of these approaches, along with other recent works such as PoseCNN \cite{xiang2017posecnn}, SSD-6D \cite{Kehl_2017_ICCV}, and BB8 \cite{Rad_2017_ICCV}, use deep convolutional neural networks (CNNs) to provide real-time, accurate pose estimation of known objects in cluttered scenes.

CNN-based pose estimation techniques enable significant improvements in the accuracy of object detection and pose estimation. However, these approaches usually require a large amount of training data containing objects of interest annotated with precise 6-DoF poses. Object poses are expensive to annotate and were often hand annotated in the past \cite{zeng2017multi,rennie2016dataset}. More recently, automatic annotation methods have been proposed using motion capture \cite{segicp} or 3D scene reconstruction \cite{lai2011large,marion2017pipeline}, but these methods still require significant human labor and are not able to generate significant variability in pose since objects must remain stationary during data capture.

To address this issue, we propose a novel pose estimation approach that leverages synthetic pose data. Our approach decouples the object pose estimation task into two cascaded components: a segmentation network and a pose interpreter network. Given an RGB image, the segmentation network first generates object segmentation masks, which are then fed into the pose interpreter network for pose estimation.

Pose interpreter networks perform 6-DoF object pose estimation on object segmentation masks and are trained entirely using synthetic pose data, thus obviating the need for expensive annotation of object poses. Using a rendering engine, we cheaply acquire large quantities of synthetic object segmentation masks and their 6-DoF pose ground truth, covering the full space of object poses. After training, our pose interpreter networks are able to estimate object pose accurately given only the segmentation mask of the object. The overall object pose estimation pipeline including segmentation runs in real-time with no postprocessing steps such as ICP refinement or smoothing.

The main contributions of this work are: (1) an end-to-end approach for real-time 6-DoF object pose estimation from RGB images, (2) a pose interpreter network for 6-DoF pose estimation in both real and synthetic images, which we train entirely on synthetic data, and (3) a novel loss function for regressing 6-DoF object pose. In the following sections, we discuss related work in Section \ref{related}, describe our technical approach in Section \ref{approach}, present experimental results in Section \ref{experiments}, and conclude in Section \ref{conclusion}.
\begin{figure}
\includegraphics[width=0.49\columnwidth]{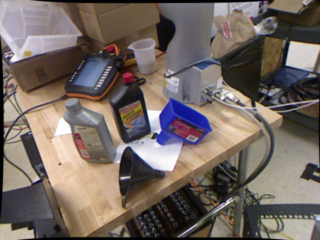}
\includegraphics[width=0.49\columnwidth]{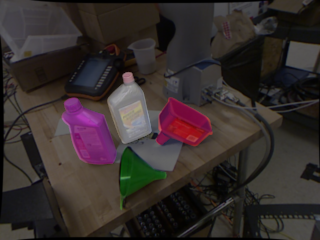}
\caption{Our end-to-end network takes in an RGB image and outputs 6-DoF object poses for all recognized objects in the image.}
\label{fig:intro}
\vspace{-10pt}
\end{figure}

\begin{figure*}
\includegraphics[width=\textwidth]{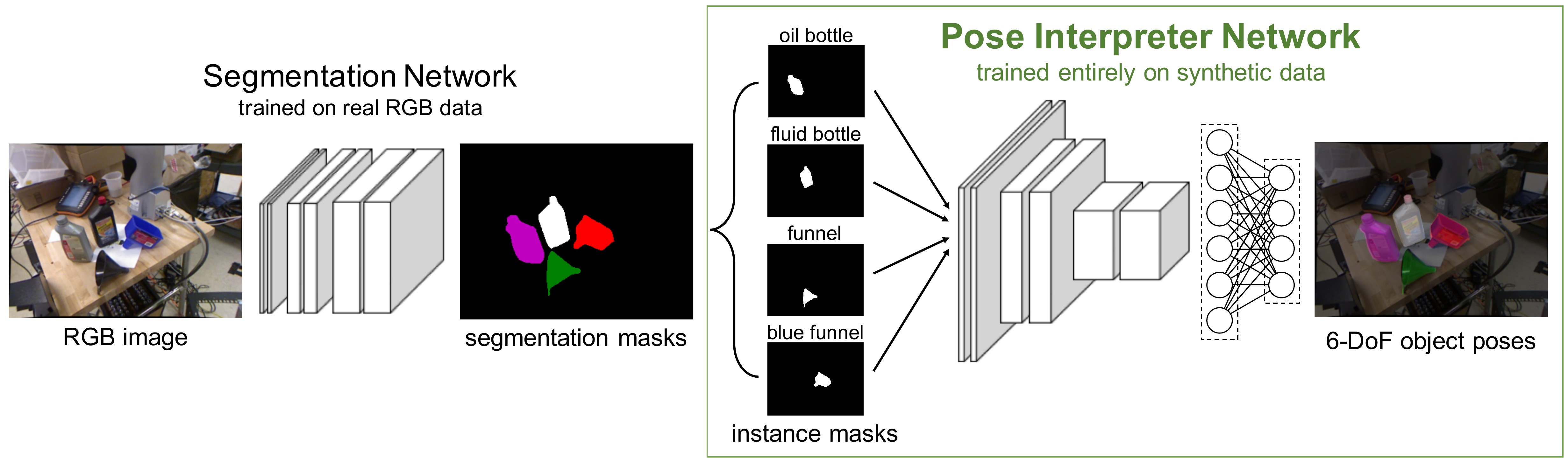}
\caption{\textbf{Full end-to-end architecture for object pose estimation on RGB images.} We use a segmentation network to extract instance masks labeled with object class, which serve as input to the pose interpreter network. The pose interpreter network operates on single object instances of known object classes and is trained entirely on synthetic object masks. The system makes one forward pass through the segmentation network for each image. Then, for each object instance, it makes one forward pass through the pose interpreter network to predict the object's pose. During evaluation, the two component networks are combined into a single end-to-end neural network.}
\label{fig:arch}
\end{figure*}

\section{Related Work}
\label{related}

Prior work in object pose estimation from RGB images include template matching approaches \cite{hinterstoisser2012model,hinterstoisser2012gradient,huttenlocher1993comparing,lowe2001local} and parts-based models \cite{savarese20073d,lim2014fpm}, which work well for highly textured objects. Feature-based methods match features in images with corresponding parts of 3D models \cite{lowe1999object,collet2011moped,rothganger20063d,pavlakos20176}. For RGB-D data, pose estimation has traditionally used variants of the iterative closest point (ICP) algorithm \cite{besl1992method,rusinkiewicz2001efficient}. More recent works have used feature matching on 3D data \cite{brachmann2014learning,wohlhart2015learning,kehl2016deep,doumanoglou2016siamese,zeng20163dmatch} or probabilistic methods \cite{Krull_2015_ICCV,Brachmann_2016_CVPR}.

With the recent successes of object recognition \cite{russakovsky2015imagenet,krizhevsky2012imagenet}, object detection \cite{girshick2014rich,liu2016ssd}, and segmentation \cite{long2015fully,Yu2017} in 2D images, many works have extended or incorporated these methods in 6-DoF pose estimation \cite{gupta2015aligning,Bansal16,Kehl_2017_ICCV,Rad_2017_ICCV}, including end-to-end systems for robotic manipulation \cite{segicp,zeng2017multi,zeng2018robotic}. In contrast to these approaches, which often require expensively obtaining lots of annotated training data, we focus on the use of cheaply acquired synthetic data to train our pose interpreter network, and show that it generalizes well to real RGB images.

Perhaps most closely related to our work is PoseCNN \cite{xiang2017posecnn}, a well-known CNN for 6-DoF object pose estimation. We emphasize that our pose interpreter network is trained entirely on synthetic data, whereas PoseCNN uses a large annotated pose dataset augmented with synthetic images. Additionally, our system runs in real-time and uses neural network forward passes to directly output pose estimates, without any further postprocessing.

Our work, which use CNNs for regression, is also related to \cite{kendall2015posenet} and \cite{Kendall_2017_CVPR}, known for successfully demonstrating camera pose regression with CNNs. Additionally, the use of rendering software to cheaply acquire large quantities of synthetic training images for training deep networks has been proposed by several previous works \cite{su2015render,wu2016single,marrnet}. In particular, \cite{wu2016single} also uses an intermediate representation to bridge synthetic data and real data for 3D object structure recovery.

\section{Technical Approach}
\label{approach}

Our approach to object pose estimation consists of a two step process. We first use a segmentation network to generate object instance masks. The masks are then passed individually through the pose interpreter network, which outputs a 6-DoF pose estimate for each object. While we train the segmentation model on real RGB images, our pose interpreter network is trained entirely on synthetic data. We first describe our segmentation network in Section \ref{segmentation}. Then, we describe the pose interpreter network in Sections \ref{pose_representation} through \ref{losses}.

\subsection{Segmentation Network}
\label{segmentation}

We use a dilated residual network (DRN) \cite{Yu2017} trained for semantic segmentation as the first component of our end-to-end system. The network takes in real RGB images and outputs segmentation labels, which are converted into binary instance masks with associated object classes and fed into the subsequent pose estimation network.

In contrast to regular residual networks \cite{he2016deep}, which use subsampling to increase receptive field size at the cost of spatial acuity, DRNs use dilated convolutions, which preserve spatial resolution while maintaining high receptive fields. As a result, these networks are particularly well suited for dense prediction tasks such as semantic segmentation. Compared to other architectures for semantic segmentation such as SegNet \cite{badrinarayanan2015segnet} or DeepLab \cite{chen2017deeplab}, we observed that DRNs trained on our RGB image dataset generated higher fidelity segmentations with fewer false positives.

While our segmentation training data is not synthetically generated, we note that compared to CNNs for pose estimation, CNNs for segmentation can use cheaper data acquisition techniques and much more aggressive data augmentation. We also note that our use of a semantic segmentation model for instance segmentation assumes that there is at most one instance of every object class. However, our system can be adapted to handle multiple instances by simply swapping out the semantic segmentation component with an instance segmentation model such as Mask R-CNN \cite{he2017mask}.

\subsection{Object Pose Representation}
\label{pose_representation}

We represent the pose of an object by its position $\textbf{p} = (x, y, z)$ and orientation $\textbf{q} = (q_0, q_x, q_y, q_z)$, which are translations and rotations relative to the camera coordinate frame. Any given rotation can have multiple equivalent forms, and we found it crucial to enforce that only a single unique form is valid. For example, the axis-angle rotation $(\omega, \theta)$ is equivalent to $(-\omega, -\theta)$, a rotation of $-\theta$ about the axis $-\omega$. These two forms resolve to a unique form in unit quaternion. The rotation $(\omega, \theta)$ is also equivalent to $(-\omega, 2\pi-\theta)$, which resolves to $-q$ in unit quaternion. We resolved this equivalence of $q$ and $-q$, known as double cover \cite{altmann2005rotations}, by requiring that the real component of the quaternion $q_0$ be nonnegative, equivalent to constraining the rotation angle $\theta$ to be in the range $(-\pi,\pi)$.

\subsection{Pose Interpreter Network}

The pose interpreter network operates on single object instances of known object classes, and is trained entirely on synthetic data. The network follows a simple CNN architecture consisting of a ResNet-18 \cite{he2016deep} feature extractor followed by a multilayer perceptron, as illustrated in Fig. \ref{fig:arch}. We removed the global average pooling layer from the feature extractor to preserve spatial information in the feature maps.

The multilayer perception is composed of one fully connected layer with 256 nodes, followed by two parallel branches corresponding to position and orientation, respectively. Each branch consists of another single fully connected layer, with a separate set of outputs for each object class. We train our pose interpreter network with five object classes, so the position branch has 15 outputs, while the orientation branch has 20.

The quaternion orientation outputs are normalized to unit magnitude. We found this normalization to be crucial, as it is difficult to directly regress unit quaternion values. By normalizing the outputs, we are instead having our network predict the relative weights of the four quaternion components.

\subsection{Point Cloud L1 Loss for Pose Prediction}
\label{losses}

We propose a new loss function for object pose prediction, the Point Cloud L1 Loss. We compare the proposed loss with several baseline loss functions and show our experimental results in Section \ref{losses_expts}.

The simplest baseline is L1 loss on the target and output poses, with a weighting constant $\alpha$ to balance the position and orientation terms:

\begin{equation}
L_1 = \left| \hat{\mathbf{p}} - \mathbf{p} \right| +\alpha \left| \hat{\mathbf{q}} - \mathbf{q} \right|
\end{equation}

We also consider a modified version of $L_1$ in which we replaced the orientation loss with one proposed by Xiang et al. in PoseCNN \cite{xiang2017posecnn}, which approximates the minimum distance between the target and predicted orientations:

\begin{equation}
L_2 = \left| \hat{\mathbf{p}} - \mathbf{p} \right| + \alpha \left( 1 - \langle \hat{\mathbf{q}}, \mathbf{q} \rangle \right)
\end{equation}

Next, we propose a new loss function, denoted $L_3$ below, that operates entirely in the 3D space rather than the quaternion space. Using the 3D models of objects in our synthetic dataset, we generate point clouds representing each object. Given a target pose and output pose for an object, we transform the object's point cloud using both the target and output poses and compare the two transformed point clouds. We compute an L1 loss between pairs of corresponding points as follows:

\begin{equation}
L_3 = \sum_{i=1}^{m}{\left|H(\hat{\mathbf{p}}, \hat{\mathbf{q}})\mathbf{x}^{(i)} - H(\mathbf{p}, \mathbf{q})\mathbf{x}^{(i)}\right|}
\end{equation}
where each $x^{(i)}$ is one of $m$ points in the object point cloud, and the function $H$ transforms an object pose $(\mathbf{p}, \mathbf{q})$ into the equivalent transformation matrix. This loss function directly compares points clouds in 3D space and does not require tuning an additional hyperparameter $\alpha$, as in $L_1$ and $L_2$, to balance the position and orientation loss components.

As discussed in Section \ref{pose_representation}, we require that the first component $q_0$ of unit quaternion orientations be nonnegative. In practice, we found that adding an additional term to penalize predictions with negative $q_0$ helped with convergence. Thus, we favor the use of loss $L_4$ below, a variant of our proposed loss $L_3$.

\begin{equation}
L_4 = \max(-q_0, 0) + \sum_{i=1}^{m}{\left|H(\hat{\mathbf{p}}, \hat{\mathbf{q}})\mathbf{x}^{(i)} - H(\mathbf{p}, \mathbf{q})\mathbf{x}^{(i)}\right|}
\end{equation}

\section{Experiments}
\label{experiments}

We present our experimental results in the following sections. Section \ref{datasets} describes in detail the two datasets we used. Section \ref{drn} describes the performance of our segmentation network, which serves as the first component of our end-to-end system. Sections \ref{pose}, \ref{losses_expts}, and \ref{quantity} describe the results of our experiments with the pose interpreter network on synthetic data. In Sections \ref{real} and \ref{live}, we combine the segmentation network and pose interpreter network into our full end-to-end system and evaluate its performance on real RGB data. Finally, in Sections \ref{limitations} and \ref{occlusion_training}, we discuss limitations of our approach as well as experiments to address some of those limitations.

\subsection{Datasets}
\label{datasets}

As shown in Fig. \ref{fig:arch}, we use separate datasets for training the segmentation network and the pose interpreter network. The segmentation network is trained on real RGB images, while the pose interpreter network is trained entirely on synthetic data. We describe each of the two datasets in detail below.

\textbf{Oil Change Dataset.} We use an extension of the dataset used in the SegICP system \cite{segicp}, which we will refer to as the Oil Change dataset. The dataset consists of indoor scenes with 10 categories of densely annotated objects relevant to an automotive oil change, such as oil bottles, funnels, and engines. Images were captured with one of three sensor types (Microsoft Kinect1, Microsoft Kinect2, or Asus Xtion Pro Live) and were automatically annotated with object poses and pixelwise instance masks using either the motion capture setup described in \cite{segicp} or the LabelFusion \cite{marion2017pipeline} pipeline.

In total, there are 7,879 images used for training and another held out 1,950 test images used for evaluation. For our segmentation model, we used the full training and testing splits. For evaluating our end-to-end system, we used a subset of the testing images corresponding to one of the Kinect1 cameras. This is because unlike in most 2D image recognition tasks, the task of recovering 3D pose from a 2D image depends on the intrinsics of the camera with which the image was taken, so we cannot use images from several different cameras. As a result, we focus on our most portable Kinect1 camera for our end-to-end system. The subset we used for evaluating the end-to-end system consisted of 229 test images with 683 object instances from 5 categories. These same 5 object categories were used for our synthetic image dataset, described in further detail below.

\begin{figure}
\includegraphics[width=\columnwidth]{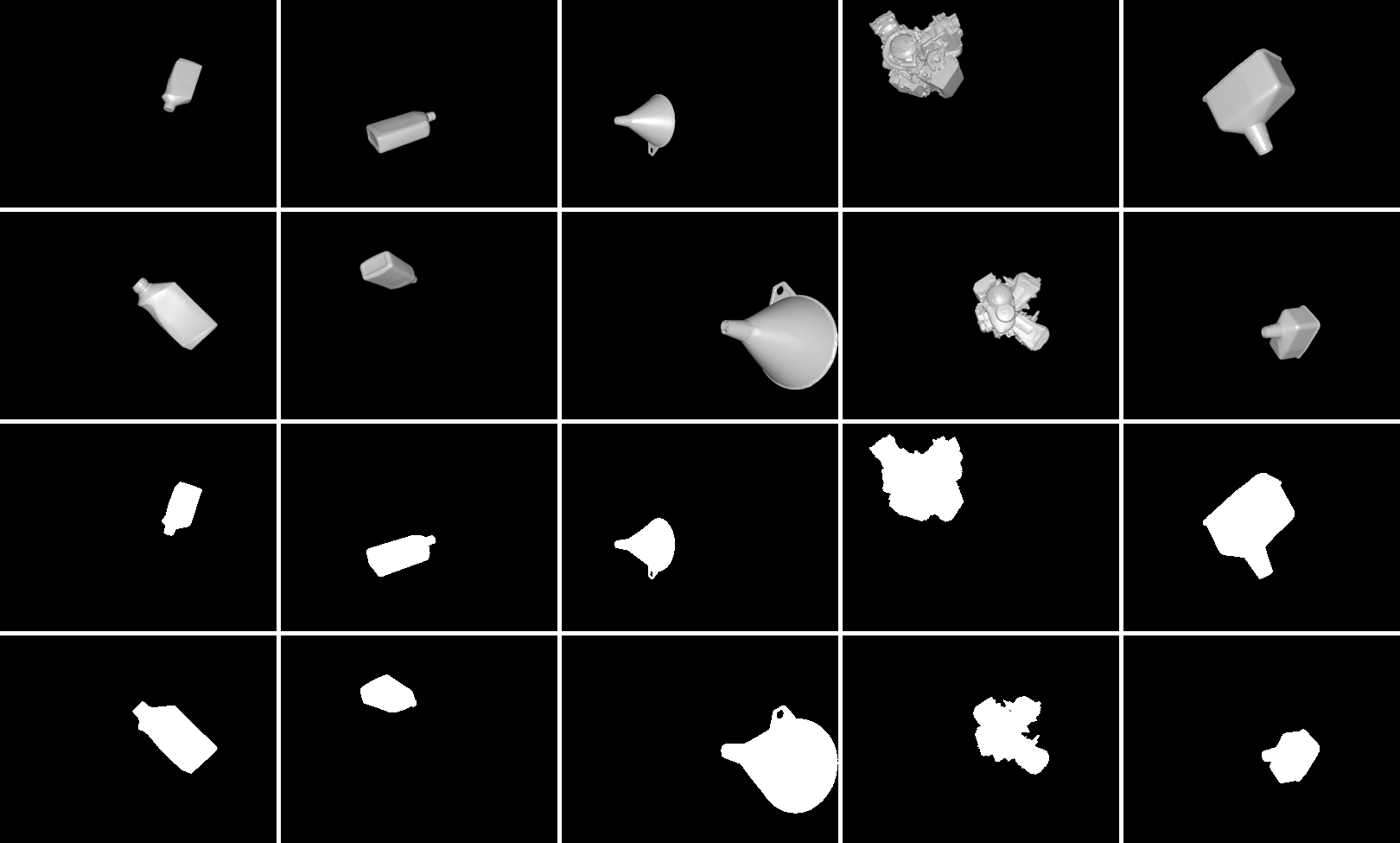}
\caption{We use a synthetic image dataset of five object classes to train our pose interpreter network. The dataset contains both synthetic object images (top two rows) and synthetic mask images (bottom two rows).}
\label{fig:synthetic}
\end{figure}

\textbf{Synthetic Image Dataset.} We train and evaluate our pose interpreter network entirely on a synthesized dataset of object images and mask images, examples of which are shown in Fig. \ref{fig:synthetic}. We load 3D model files for five object categories from the Oil Change dataset and use the Blender rendering software to render the objects in random poses. As previously discussed, we focus on one Kinect1 camera for evaluating our end-to-end system. In order to ensure compatibility with the evaluation images, we calibrated our camera and used the same camera intrinsics in the rendering pipeline.

While it is possible to render images on the fly during neural network training, we found that the rendering time far exceeded the network training step time. We thus rendered and saved a total of 3.2 million training images and 3,200 testing images, evenly spanning the five object classes. We note that much less training data is actually needed, and in Section \ref{quantity}, we investigate the effect of training data quantity on the performance of the pose interpreter network.

\subsection{Semantic Segmentation with DRN}
\label{drn}

Our segmentation model is a DRN-D-22 trained on the Oil Change dataset with batch size 16, learning rate 0.001, momentum 0.99, and weight decay 0.0001. The dataset was annotated with 11 classes, corresponding to the 10 object classes and an additional background class. We trained for 900 epochs and used aggressive data augmentation to improve generalization, including random scaling, random rotations, random cropping, and gamma jittering. On the Oil Change test images, our final model achieves a pixelwise accuracy of 99.82\% and a mean IoU of 0.9650 across the 10 object classes.

\subsection{Pose Interpreter Network}
\label{pose}

As described in Section \ref{datasets} and Fig. \ref{fig:synthetic}, we used a synthetic dataset containing both object images and mask images. We use the synthetic object images for further experiments with the pose interpreter network. The synthetic mask images are used to train the pose interpreter network for use in our full end-to-end system.

In Table \ref{tab:pose}, we show the performance of our pose interpreter network after training and evaluating on the synthetic dataset for both object images and mask images. We use a batch size of 32, weight decay of 0.0001, and train for 21 epochs with an initial learning rate of 0.01, which we decay by a factor of 10 after 7 epochs and 14 epochs. A single network is trained to handle all five object classes with separate output heads per class. By design, the network operates on instance masks of known object class, so for each training example we are able to select the appropriate outputs to compute the loss on.

As one might expect, the model trained on synthetic mask images does not perform as well as the model trained on synthetic object images, as there is less information and more ambiguity when given only a binary mask of an object. However, the model trained on synthetic mask images can be directly used on real RGB data when combined with a segmentation model. We describe the results of this combination applied to real RGB data in Section \ref{real}.

\begin{table}
\caption{Performance of the pose interpreter network trained and evaluated on synthetic data.}
\label{tab:pose}
\begin{center}
\begin{tabular}{ c c c } 
\hline
model type & pos. error (cm) & ori. error (deg) \\ 
\hline
object & 1.12 & 8.93 \\ 
mask & 1.43 & 14.83 \\
\hline
\end{tabular}
\end{center}
\vspace{-10pt}
\end{table}

\subsection{Pose Prediction Loss Functions}
\label{losses_expts}

We considered four loss functions and evaluated them on a subset of the synthetic dataset consisting of synthetic blue funnel object images. We used an initial learning rate of 0.01 for all four training runs, which was decayed after observing the validation performance plateau (10 epochs for $L_1$, 30 epochs for $L_2$, and 15 epochs for $L_3$ and $L_4$). The weighting term $\alpha$ was set to $1$ for $L_1$ and $L_2$. We show in Table \ref{tab:loss} the best performance for each network after 30 epochs (45 for $L_2$ due to slower convergence).

We observed that the network trained using $L_1$ attains comparable performance to the point cloud loss functions $L_3$ and $L_4$. However, the weighting term $\alpha$ for $L_1$ must be tuned to balance the position and orientation errors. Lowering the position error by adjusting the weighting term would raise the orientation error, and vice versa. By contrast, our proposed point cloud loss function naturally balances the position and orientation errors by computing the loss in the 3D point space, so there is no weighting term to tune.

\begin{table}
\caption{Performance of pose interpreter network trained with various loss functions on synthetic blue funnel object images. The loss functions are described in Section \ref{losses}.}
\label{tab:loss}
\begin{center}
\begin{tabular}{ c c c } 
\hline
loss function & pos. error (cm) & ori. error (deg) \\ 
\hline
$L_1$ & 0.75 & 5.79 \\ 
$L_2$ & 1.38 & 13.76 \\
$L_3$ & 0.58 & 6.25 \\ 
$L_4$ & 0.50 & 6.01 \\
\hline
\end{tabular}
\end{center}
\end{table}

\subsection{Synthetic Training Data Quantity}
\label{quantity}
We used 3.2 million training images to train our pose interpreter network. Here we investigate whether comparable performance can be attained with fewer training images, and whether the training scales well with more object classes. We run two sets of experiments using subsets of our synthetic dataset, the first on synthetic object images of a single object class (blue funnel) as shown in Table \ref{tab:blue_funnel_quantity}, and the second on synthetic object images of all five object classes as shown in Table \ref{tab:quantity}. We used a learning rate of 0.01 with no decay for all experiments, and show the best performance attained after an equal number of training iterations (400k for single object class, 700k for five object classes).

The results in Table \ref{tab:blue_funnel_quantity} indicate that when training the pose interpreter network on a single object class, using more than 100k training images does not yield further gains. When training the pose interpreter network on multiple object classes instead, the results in Table \ref{tab:quantity} indicate that 25k images is the cutoff point. We interpret this as evidence that the network is learning some shared knowledge between the different object classes, as only 25k images per class are needed rather than 100k. Although more training images and training iterations are required for multiple object classes, the required quantities are far from proportional to the number of classes.

\begin{table}
\begin{center}
\caption{Performance of pose interpreter network trained on synthetic object images from the blue funnel object class. We vary the number of training images used to see the effect on performance.}
\label{tab:blue_funnel_quantity}
\begin{tabular}{ c c c } 
\hline
images & pos. error (cm) & ori. error (deg) \\ 
\hline
12,800 & 3.16 & 104.6 \\ 
25,600 & 1.86 & 48.15 \\
51,200 & 1.50 & 11.85 \\ 
102,400 & 1.06 & 10.99 \\
640,000 & 1.33 & 11.02 \\
\hline
\end{tabular}
\end{center}
\end{table}

\begin{table}
\begin{center}
\caption{Performance of pose interpreter network trained on synthetic object images from all five object classes. We vary the number of training images used per object class.}
\label{tab:quantity}
\begin{tabular}{ c c c } 
\hline
images / class & pos. error (cm) & ori. error (deg) \\ 
\hline
6,400 & 4.91 & 82.45 \\
12,800 & 3.36 & 27.06 \\
25,600 & 3.21 & 22.86 \\
51,200 & 2.94 & 25.16 \\ 
640,000 & 2.96 & 24.26 \\
\hline
\end{tabular}
\end{center}
\end{table}

\begin{figure*}
\includegraphics[width=0.49\textwidth]{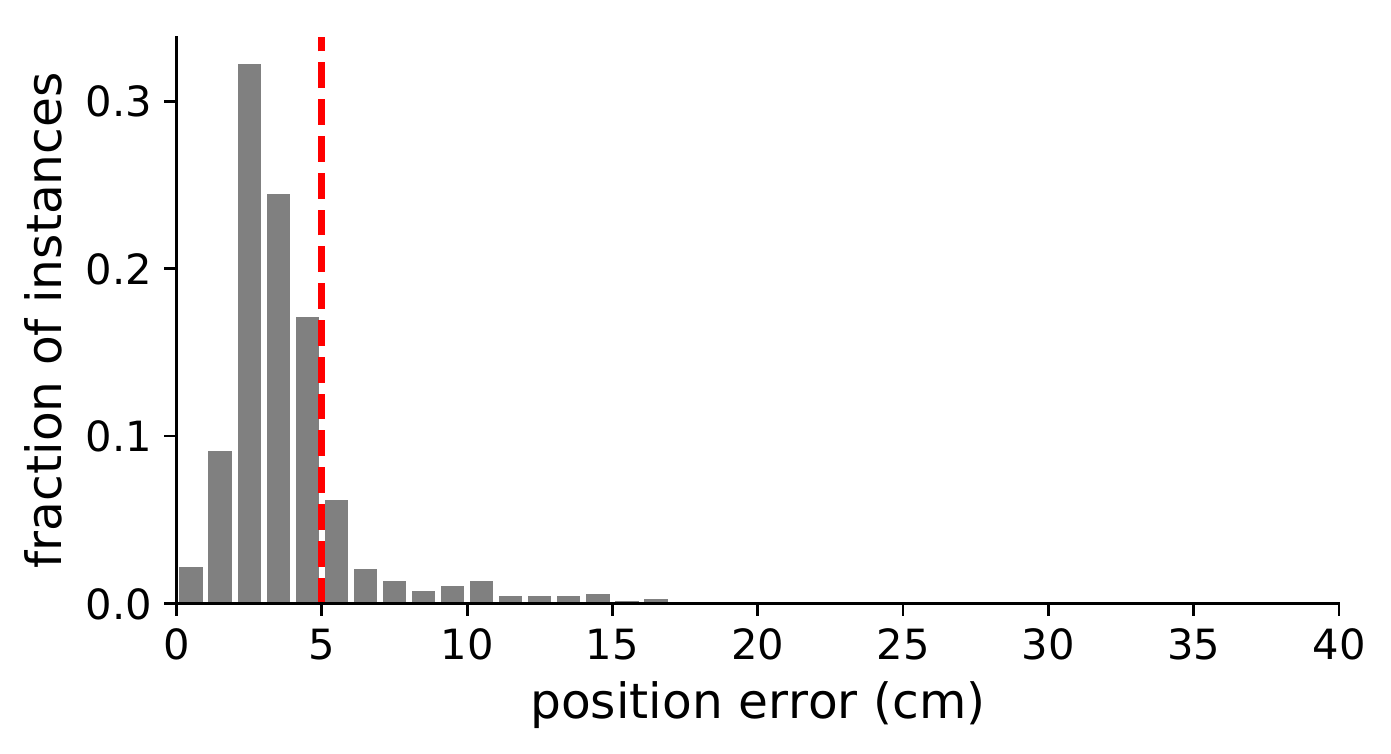}
\includegraphics[width=0.49\textwidth]{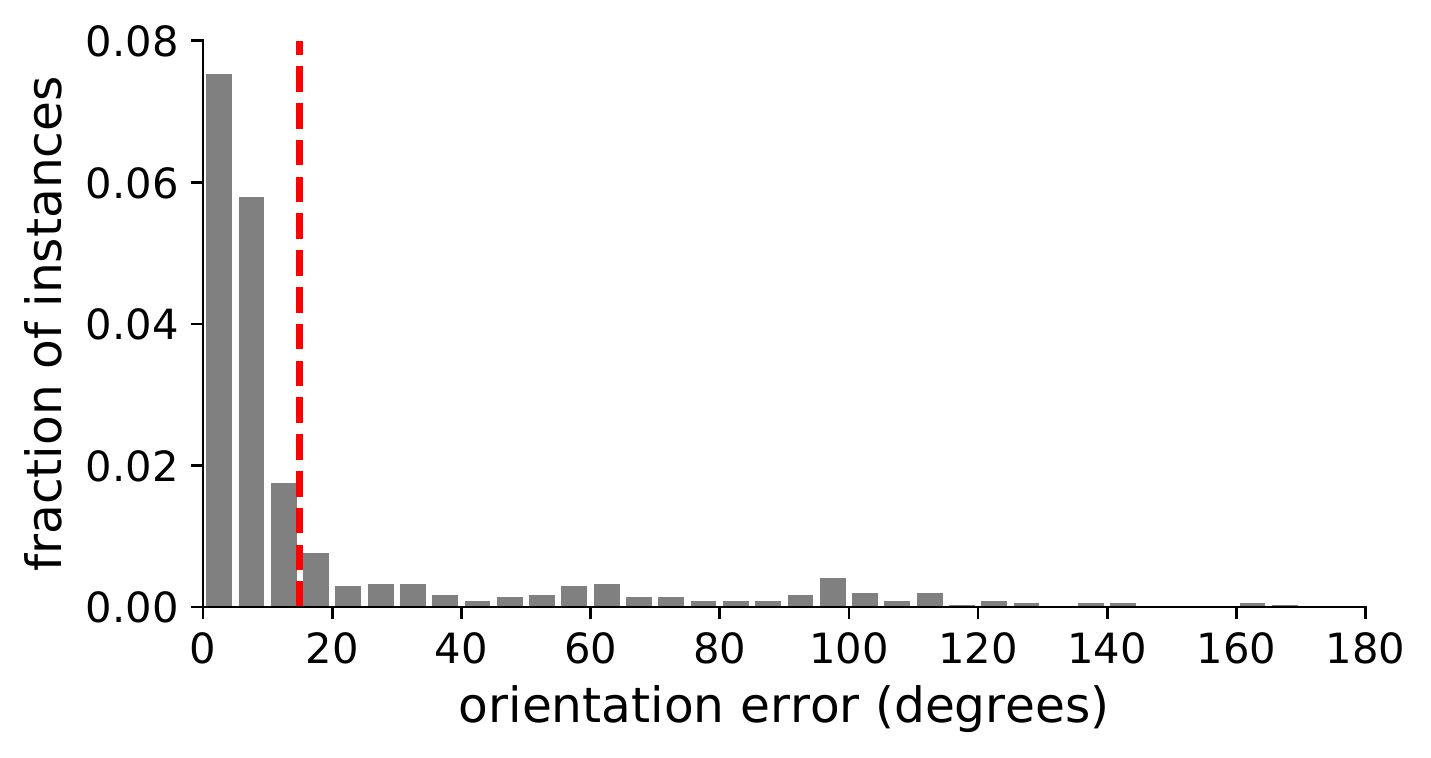}
\caption{Histograms showing position errors and orientation errors of our end-to-end pose estimation system on real RGB data. The success criterion of $< 5~\text{cm}$ position error and $< 15^\circ$ orientation error is indicated by the dotted red lines. Our system attains a success rate of 71.01\%.}
\label{fig:error_histograms}
\end{figure*}

\subsection{Object Pose Estimation on Real RGB Images}
\label{real}

We evaluate our end-to-end object pose estimation system on real RGB images from the Oil Change dataset test split. The end-to-end system is composed of a DRN segmentation model followed by a object mask pose interpreter network. As described in Section \ref{datasets}, we evaluate only on the images taken with a specific Kinect1 sensor, consisting of 229 images with 683 total object instances.

\begin{table}
\caption{End-to-end performance of our object pose estimation system on real RGB images. We also show the performance for SegICP evaluated on the same test images.}
\label{tab:full_perf}
\begin{center}
\begin{tabular}{ c c c c } 
\hline
& & ours & SegICP \\
\hline
\hline
pos. error (cm) & mean & 3.76 & 2.09 \\
& median & 3.23 & 1.32 \\
\hline
ori. error (deg) & mean & 19.64 & 68.93 \\
& median & 6.17 & 55.63 \\
\hline
success (\%) & & 71.01 & 42.08 \\
\hline
\end{tabular}
\end{center}
\end{table}

We show the performance of our end-to-end system evaluated on our Oil Change dataset in Table \ref{tab:full_perf}. Following the convention used in \cite{zeng2017multi} and \cite{segicp}, a successful pose estimate is defined as $< 5~\text{cm}$ position error and $< 15^\circ$ orientation error. Histograms of the position and orientation errors in Fig. \ref{fig:error_histograms} show the distribution of errors relative to the success cutoffs, marked in red.

We also show in Table \ref{tab:full_perf} a comparison with SegICP, which reported a success rate of 77\% in \cite{segicp}. However, that success rate was evaluated on an older version of our dataset. For a fair comparison, we evaluated the performance of SegICP on the updated test set, which proved to be more challenging than the set used in the SegICP paper. The current test set includes many object instances that lie in close proximity to other objects. This means that mistakes in the segmentation can result in the inclusion of erroneous points from neighboring objects, leading to poor ICP performance in SegICP. Furthermore, although SegICP attains better position errors than our system, we would like to emphasize that SegICP uses ICP to iteratively refine predictions, whereas our system outputs a one-shot prediction per object with no postprocessing or refinement.

\subsection{Object Pose Estimation on Live RGB Data}
\label{live}

In order to verify that our approach does not suffer from overfitting or dataset bias, we tested the efficacy of our approach on live RGB data. Our experimental setup used an Ubuntu 16 workstation equipped with a dedicated Nvidia Titan Xp graphics card. A ROS node processes live images from our Kinect1 and runs them through trained PyTorch models to generate 6-DoF pose predictions for each recognized object. We benchmark the processing time per frame by averaging over 50 consecutive frames. The DRN forward pass takes 28.9 ms on 640$\times$480 RGB images. The pose interpreter network inputs 320$\times$240 binary instance masks and requires a 3.6 ms forward pass per detected object instance. Visualization adds an additional overhead of about 10 ms. Overall, the end-to-end system takes about 32--47 ms per frame depending on the number of objects in the image. In the accompanying video supplement, we demonstrate our system performing pose estimation in real-time on live data.

\subsection{Limitations}
\label{limitations}

The main limitations of our approach are sensitivities to both occlusion and segmentation failures. Our pose interpreter network is trained entirely on synthetic data, and thus only generalizes well when input object masks resemble the rendered masks seen during training. Hence, the performance of our end-to-end system is closely tied to the quality of the segmentation model.

\begin{figure}
\includegraphics[width=\columnwidth]{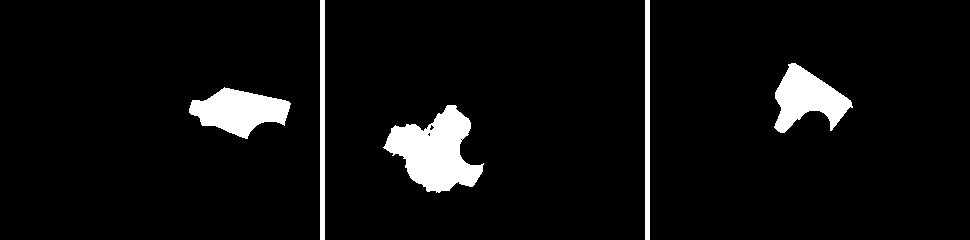}
\caption{Some examples of the artificially occluded mask images we generated. These images were used to quantify the pose interpreter network's sensitivity to occlusions. We also ran experiments, as described in Section \ref{occlusion_training}, using occluded images as training data.}
\label{fig:occlusion_gallery}
\end{figure}

\begin{figure}
\includegraphics[width=0.49\columnwidth]{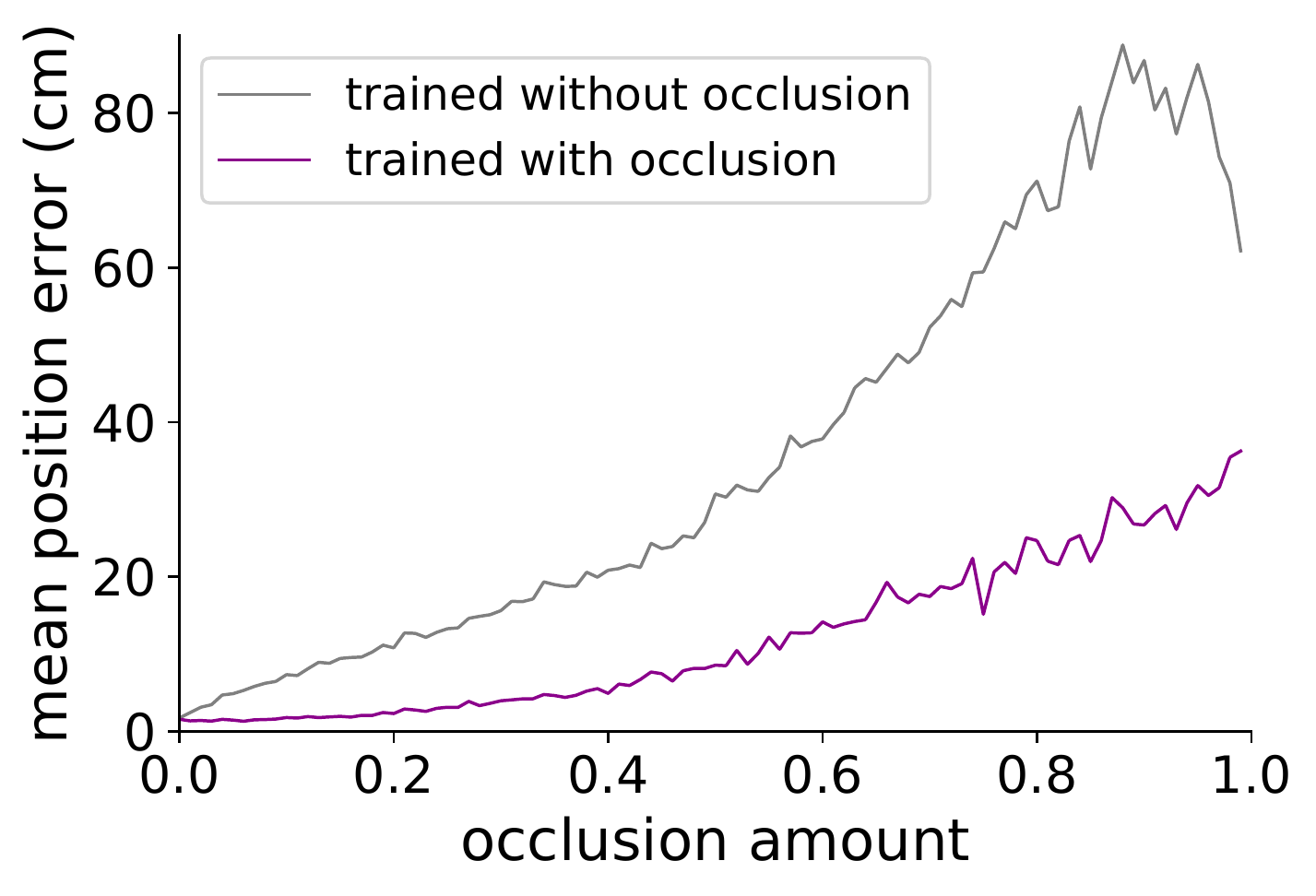}
\includegraphics[width=0.49\columnwidth]{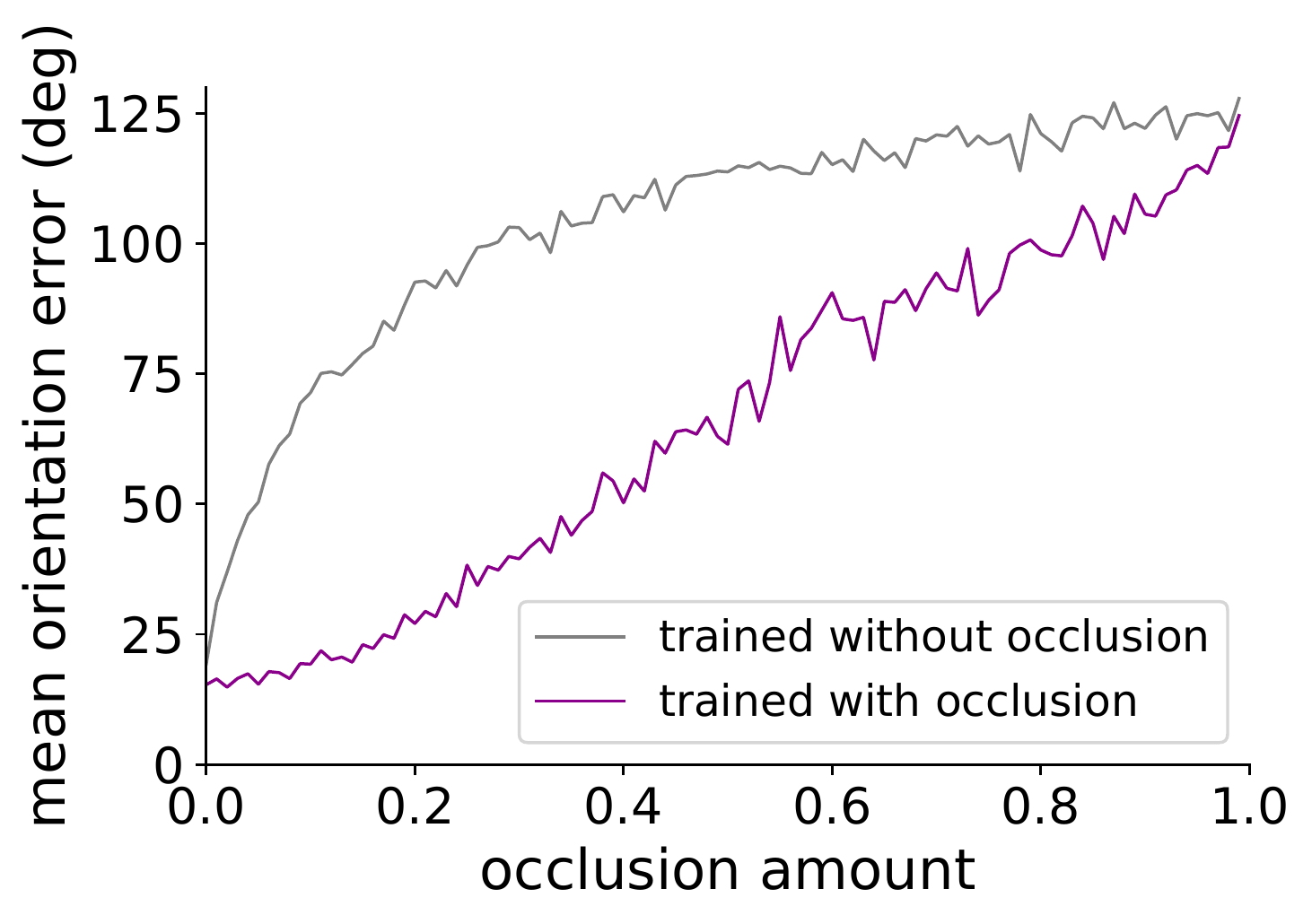}
\caption{We artificially introduce circular occlusions of various sizes, as shown in Fig. \ref{fig:occlusion_gallery}, and quantify how the performance of the object mask pose interpreter network varies with the amount of occlusion. We then train a model with occluded mask images, as detailed in Section \ref{occlusion_training}, and compare the resulting performance with the baseline model trained without occlusion.}
\label{fig:occlusion}
\end{figure}

We quantitatively evaluate the effect of occlusions on the performance of the object mask pose interpreter network by introducing circular occlusions of various sizes centered at points on mask boundaries, as shown in Fig. \ref{fig:occlusion_gallery}. We show in Fig. \ref{fig:occlusion} how the position and orientation errors of the network relate to the amount of occlusion introduced. The occlusion amount measures the fraction of the original mask area that has been artificially occluded. The results (trained without occlusion) indicate high sensitivity to occlusion, particularly for orientation prediction at low occlusion amounts. In Section \ref{occlusion_training}, we investigate whether training with artificially occluded mask images improves the robustness of the pose interpreter network.

Another limitation of our approach is the lack of additional information such as texture, color, or depth in our binary mask representation, which presents difficulties in situations where  such information is needed to resolve ambiguities in the mask representation. However, extending to other domains such as depth would require either more realistic rendering or domain adaptation techniques. Here, we would like to emphasize that our object mask representation enables us to directly apply our pose interpreter network to real RGB data without any domain adaptation.

\subsection{Training with Occlusion}
\label{occlusion_training}

As discussed in Section \ref{limitations}, our model is not robust to occlusions, particularly for orientation prediction. We investigate here whether we can improve the robustness of object mask pose interpreter networks by training on occluded mask images. Using the same occlusion scheme as in Section \ref{limitations}, we artificially occlude mask images during training and evaluate the resulting performance.

\begin{figure}
\includegraphics[width=\columnwidth]{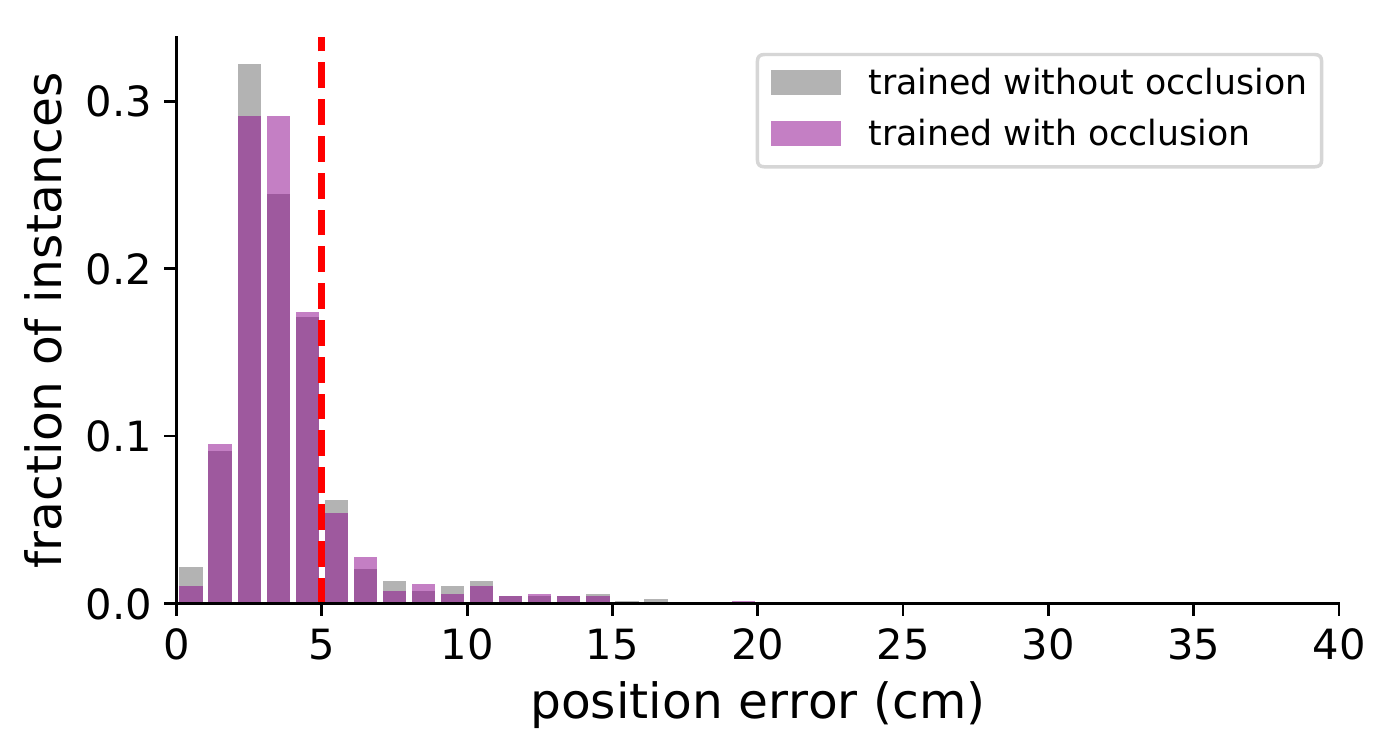}
\includegraphics[width=\columnwidth]{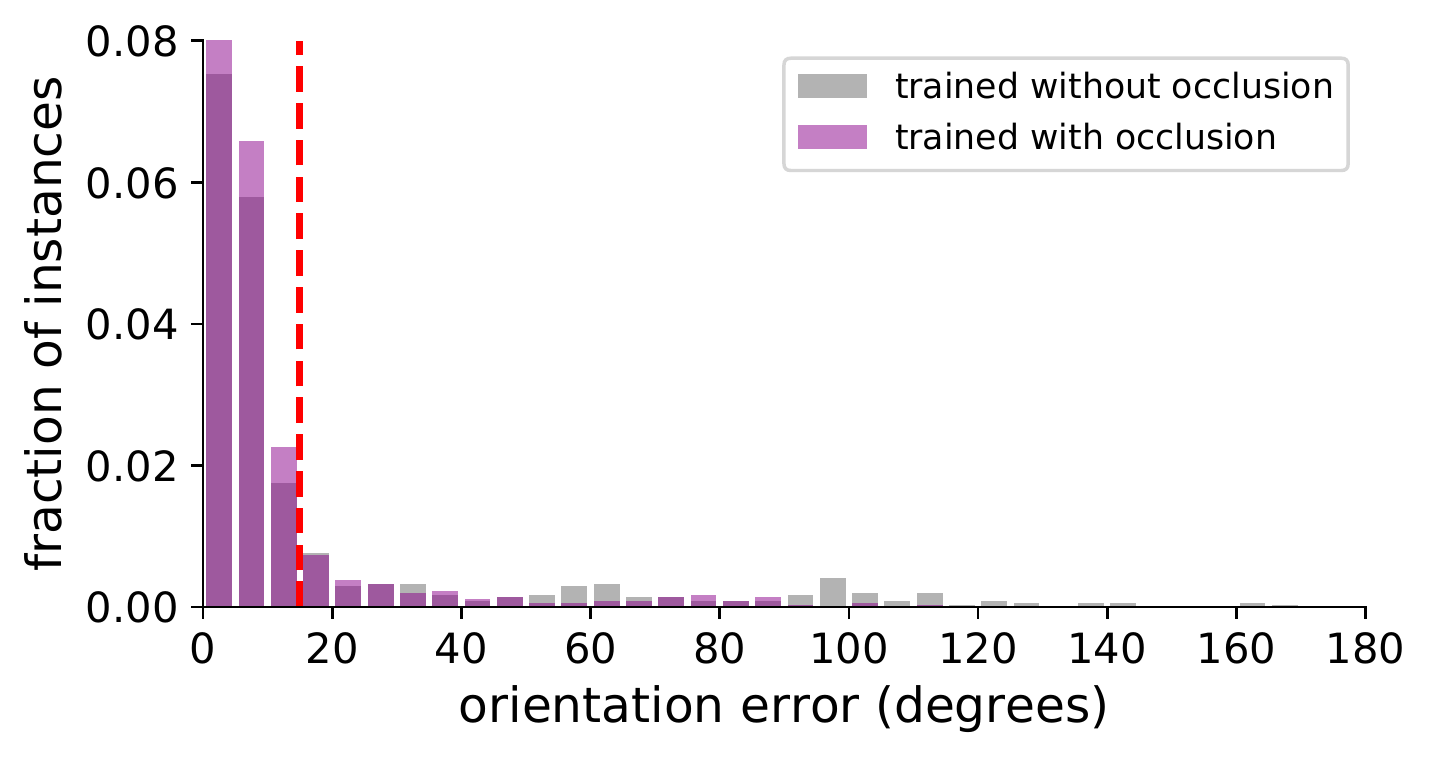}
\caption{Histograms showing position errors and orientation errors of the end-to-end system when the pose interpreter network is trained with and without occlusion. The end-to-end system is evaluated on real RGB data. For the model trained with occlusion, note that the orientation errors are much more concentrated in the low error region.}
\label{fig:error_histogram_with_occlusion}
\end{figure}

We show in Figs. \ref{fig:occlusion} and \ref{fig:error_histogram_with_occlusion} comparisons between the baseline mask model and a model trained with circular occlusions of maximum radius 24. Fig. \ref{fig:occlusion} indicates that the pose interpreter network is more robust when trained with occlusion, especially for orientation prediction at low occlusion amounts. In Fig. \ref{fig:error_histogram_with_occlusion}, we show that when applied to the end-to-end system and evaluated on real RGB images, training with occlusion does not show meaningful improvement in position prediction, but does markedly improve orientation prediction, allowing the system to successfully address many examples that previously gave high orientation errors.

We note that introducing occlusions that cover too much of the original object mask will destroy the information present in the mask, so we experimented with different settings of the maximum occlusion radius. As shown in Table \ref{tab:end_to_end_with_occlusion}, we found 24 pixels to be optimal, and confirmed that training with occlusions that were too large resulted in worsened performance. We additionally found that larger occlusions generally resulted in longer training times since the task was more difficult. While 21 epochs was sufficient for the baseline model, we trained models with larger occlusions for over 100 epochs.

% \begin{table}
% \begin{center}
% \caption{Performance of pose interpreter network when trained with occluded object masks and evaluated on non-occluded masks.}
% \label{tab:pose_estimation_with_occlusion}
% \begin{tabular}{ c c c } 
% \hline
% max occlusion radius & pos. error (cm) & ori. error (deg) \\ 
% \hline
% baseline & 1.43 & 14.83 \\ 
% 12 pixels & 1.41 & 13.87 \\ 
% 16 pixels & 1.50 & 14.95 \\
% 24 pixels & 1.51 & 15.82 \\ 
% 32 pixels & 1.49 & 15.22 \\
% 48 pixels & 2.11 & 16.21 \\
% 64 pixels & 1.87 & 17.87 \\
% \hline
% \end{tabular}
% \end{center}
% \end{table}

\begin{table}
\begin{center}
\caption{Performance of our end-to-end model when the pose interpreter network is trained with occlusion. We vary the maximum radius of the circular occlusions.}
\label{tab:end_to_end_with_occlusion}
\begin{tabular}{*6c} 
\hline
max occ. radius &  \multicolumn{2}{c}{pos. error (cm)} & \multicolumn{2}{c}{ori. error (deg)} & success (\%) \\
& mean & median & mean & median \\
\hline
baseline & 3.76 & 3.23 & 19.64 & 6.17 & 71.01 \\
12 pixels & 3.94 & 3.46 & 15.79 & 5.89 & 75.40 \\
16 pixels & 3.90 & 3.30 & 15.79 & 5.67 & 73.06 \\
24 pixels & 3.76 & 3.31 & 11.55 & 5.90 & 77.89 \\
32 pixels & 4.00 & 3.67 & 13.69 & 6.92 & 72.18 \\
48 pixels & 2.84 & 2.32 & 16.88 & 7.05 & 74.38 \\
64 pixels & 4.22 & 3.88 & 19.69 & 7.78 & 59.74 \\
\hline
\end{tabular}
\end{center}
\end{table}

\section{Conclusion}
\label{conclusion}

In this work, we present pose interpreter networks for real-time 6-DoF object pose estimation. Pose interpreter networks are trained entirely using cheaply rendered synthetic data, allowing us to avoid expensive annotation of large pose datasets. We use pose interpreter networks as part of an end-to-end system for pose estimation in real RGB images. The system consists of two steps: (1) a segmentation network to generate object instance masks, and (2) a pose interpreter network which takes in instance masks and outputs pose estimates. We use the object mask as a context-independent intermediate representation that allows the pose interpreter network, trained only on synthetic data, to also work on real data. Our end-to-end system runs in real-time on live RGB data, and does not use any filtering or postprocessing to refine its pose estimates.

% \begin{figure*}
% \includegraphics[width=\textwidth]{figures/unit_gallery}
% \caption{Example unit visualizations of the pose interpreter network trained on synthetic object images. Each row shows the 8 highest activating images for one unit in the last convolutional layer. Units typically learned to detect parts of objects. From top to bottom, we see that the shown units detect insides of funnels, clutch covers of engines, bottle caps of oil bottles, and tabs of funnels.}
% \label{fig:visualization}
% \end{figure*}


%%%%%%%%%%%%%%%%%%%%%%%%%%%%%%%%%%%%%%%%%%%%%%%%%%%%%%%%%%%%%%%%%%%%%%%%%%%%%%%%

%%%%%%%%%%%%%%%%%%%%%%%%%%%%%%%%%%%%%%%%%%%%%%%%%%%%%%%%%%%%%%%%%%%%%%%%%%%%%%%%



% \section*{ACKNOWLEDGMENT}
\section*{Acknowledgement}

We thank Lucas Manuelli, Russ Tedrake, Leslie Pack Kaelbling, Tom{\'a}s Lozano-P{\'e}rez, and Scott Kuindersma for their insight and feedback. 

% The preferred spelling of the word ÒacknowledgmentÓ in America is without an ÒeÓ after the ÒgÓ. Avoid the stilted expression, ÒOne of us (R. B. G.) thanks . . .Ó  Instead, try ÒR. B. G. thanksÓ. Put sponsor acknowledgments in the unnumbered footnote on the first page.

%%%%%%%%%%%%%%%%%%%%%%%%%%%%%%%%%%%%%%%%%%%%%%%%%%%%%%%%%%%%%%%%%%%%%%%%%%%%%%%%
\bibliographystyle{IEEEtran}
\bibliography{references}

\end{document}